# Aneumo: A Large-Scale Comprehensive Synthetic Dataset of Aneurysm Hemodynamics

Xigui Li[1,2], Yuanye Zhou[2], Feiyang Xiao[1,2], Xin Guo[2,*], Yichi Zhang[1,2], Chen Jiang[2], Jianchao Ge[2], Xiansheng Wang[2], Qimeng Wang[2], Taiwei, Zhang[2], Chensen Lin[1,2,*], Yuan Cheng[1,2,*], Yuan Qi[1,2]

[1]Artificial Intelligence Innovation and Incubation Institute, Fudan University, Shanghai, China. [2]Shanghai Academy of Artificial Intelligence for Science, Shanghai, China. [*]Corresponding authors: guoxin@sais.com.cn, linchensen@fudan.edu.cn, cheng_yuan@fudan.edu.cn

**Intracranial aneurysm (IA) is a common cerebrovascular disease that is usually asymptomatic but may cause severe subarachnoid hemorrhage (SAH) if ruptured. Although clinical practice is usually based on individual factors and morphological features of the aneurysm, its pathophysiology and hemodynamic mechanisms remain controversial. To address the limitations of current research, this study constructed a comprehensive hemodynamic dataset of intracranial aneurysms. The dataset is based on 466 real aneurysm models, and 10,000 synthetic models were generated by resection and deformation operations, including 466 aneurysm-free models and 9,534 deformed aneurysm models. The dataset also provides medical image-like segmentation mask files to support insightful analysis. In addition, the dataset contains hemodynamic data measured at eight steady-state flow rates (0.001 to 0.004 kg/s), including critical parameters such as flow velocity, pressure, and wall shear stress, providing a valuable resource for investigating aneurysm pathogenesis and clinical prediction. This dataset will help advance the understanding of the pathologic features and hemodynamic mechanisms of intracranial aneurysms and support in-depth research in related fields. Dataset hosted at https://github.com/Xigui-Li/Aneumo.**

## Background & Summary

Intracranial aneurysm is a common cerebrovascular disease that is usually asymptomatic and affects about two to five percent of the world's population. The main risk is that the wall of the diseased aneurysm may suddenly rupture, resulting in severe subarachnoid hemorrhage[1]. Researchers have long been concerned with the formation, development, and rupture of aneurysms and their stabilization stages[2]. Aneurysm formation and rupture are associated with a variety of factors, and many studies have focused on individual patient factors and morphological characteristics of intracranial aneurysms, such as gender, age, hypertension, smoking, alcohol consumption, and history of cerebrovascular disease, which are widely used in clinical practice and have appeared in international expert consensus and management guidelines such as the European Stroke Organization's guidelines for the management of intracranial aneurysms and SAH[3-5].

Despite this progress, the pathophysiologic and hemodynamic mechanisms of aneurysms remain controversial compared with the known individual factors, mainly due to their complexity and the lack of systematic nature of the relevant studies, which has resulted in the fact that these mechanisms have not yet been able to effectively guide clinical practice[6,7]. Nevertheless, the relationship between hemodynamics and the risk of intracranial aneurysm rupture has been a hot topic of research in recent years. A recent study analyzed 3804 publications on risk factors for rupture of intracranial aneurysms between 2006 and 2023 and found that there has been a significant increase in recent years in studies addressing the natural history of IA rupture and its hemodynamic risk factors, with a particular focus on the role of hemodynamic features in IA rupture and the importance of its prevention[8].

Currently, most available intracranial aneurysm datasets are primarily based on imaging data, while 3D modeling data remain relatively scarce, particularly those that include computational fluid dynamics (CFD) data. For instance, the CADA[9], CHUV[10], and ADAM[11] datasets provide imaging data of aneurysms but lack corresponding 3D models and hemodynamic information. Conversely, the AneuX[12] and IntrA[13] datasets include 3D models of aneurysms but omit imaging data and hemodynamic parameters. Thus, comprehensive datasets that integrate imaging, 3D modeling, and hemodynamic parameters are crucial for advancing our understanding of the pathogenesis of intracranial aneurysms. Notably, the CMHA[14] dataset offers cerebrovascular Computed Tomography Angiography (CTA) images and hemodynamic data from 99 patients. However, its hemodynamic parameters (e.g., mean blood flow during the cardiac cycle) are restricted to a single steady-state blood flow condition. While such single steady-state flow analysis provides fundamental data to support research, it remains insufficient for capturing the complex flow environment and dynamic physiological characteristics of intracranial aneurysms. To enable more comprehensive and in-depth physiological studies, data encompassing various steady-state blood flow rates and multiple physiological conditions are required to fully characterize blood flow behavior and its biomechanical properties within aneurysms.

In parallel, advancements in Artificial Intelligence (AI) have introduced new opportunities for intracranial aneurysm research. Currently, AI is predominantly applied to tasks such as medical image classification and rupture risk prediction. However, challenges remain in improving the accuracy of risk assessments and enhancing model interpretability. Integrating large hemodynamic datasets derived from CFD simulations with AI models has the potential to address these limitations. On one hand, hemodynamic parameters provide AI models with richer physiological information, enabling them to better understand the mechanisms underlying aneurysm formation and rupture, thereby improving the accuracy and interpretability of predictions[15]. On the other hand, AI technologies can enhance the computational efficiency of CFD simulations[16,17], accelerating the analysis of complex flow fields[18,19] and facilitating large-scale intracranial aneurysm research. This synergistic integration of CFD and AI represents a promising pathway for advancing both our understanding and management of intracranial aneurysms.

To fill the gap in existing studies, we constructed a comprehensive aneurysm hemodynamic dataset. The dataset is based on 466 real aneurysm cases in the AneuX[12] dataset, with synthetic models generated by aneurysm resection and deformation operations, totaling 10,000 3D synthetic models (466 aneurysm-free and 9,534 deformed aneurysm models) covering aneurysms of different sizes, locations, and morphologies. In addition, the dataset provides 10,466 (real and synthetic) segmentation mask files resembling medical images and contains 80,000 hemodynamic parameters, including velocity, pressure, and wall shear stress, measured under 8 steady state flows (0.001 to 0.004 kg/s). The dataset will advance the study of intracranial aneurysm hemodynamics, particularly in the area of data-driven modeling and analysis, providing a valuable resource for accurate simulation and prediction of aneurysms. Researchers can use the dataset for hemodynamic modeling, optimization, and prediction to explore the relationship between aneurysm morphology, flow characteristics, and rupture risk, leading to improved clinical diagnosis and treatment decisions.

**Methods**

**Procedure**

This study simulates the hemodynamic characteristics of an aneurysm using a 3D model derived from a real intracranial aneurysm in the AneuX[12] dataset. The study involves operations such as aneurysm-free and deformation of the 3D model, followed by numerical simulations to calculate key hemodynamic parameters, including pressure, velocity, and wall shear stress (WSS). The complete flow chart of the data processing steps is shown in Fig. 1, illustrating the detailed procedures for data preprocessing and simulation operations.

The entire study process was divided into four primary steps: (1) 3D model deformation; (2) 3D model conversion to segmentation mask; (3) mesh generation; and (4) boundary conditions definition and hemodynamic simulation. All procedures were performed by two research assistants, who received relevant training under the supervision of an expert in computational fluid dynamics. A detailed description of each step is provided below.

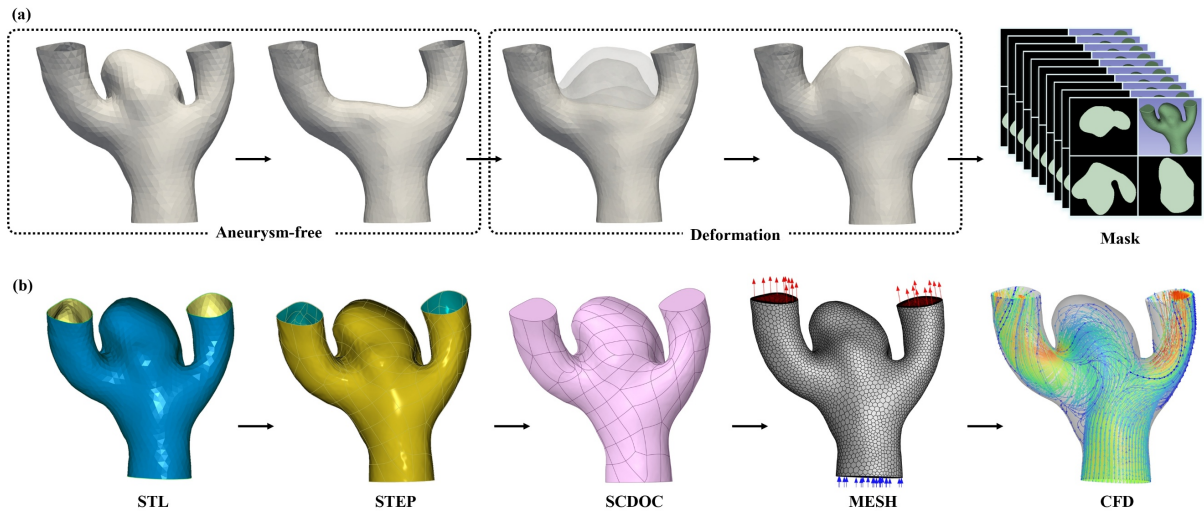

Fig. 1 Workflow of 3D aneurysm modeling, deformation, and hemodynamic simulation.

**Data analysis**

To systematically analyze the effect of geometric deformation on volume change, this study presents the distribution of the volume change rate after deformation. The volume change rate is defined as $\frac{\text{Volume}_{\text{deformed}} - \text{Volume}_{\text{aneurysm-free}}}{\text{Volume}_{\text{aneurysm-free}}}$. As shown in Fig. 2, the volume change rate of most deformed geometric models lies within the range of 0 to 1, with only a few samples falling outside this interval. This outcome aligns with expectations, indicating that within a reasonable deformation range, the geometric volume changes remain within anticipated physical limits. Notably, some models exhibit a volume change rate of less than 0. This phenomenon is primarily attributed to localized depressions near the deformed regions during the initial tensile deformation stage, which reduces the overall volume, leading to negative values for the volume change rate. Additionally, a small subset of models shows a volume change rate greater than 1.0. This is likely due to significant volume expansion in the aneurysm region caused by multiple tensile deformations, resulting in volumes far exceeding those of the aneurysm-free models. In summary, the distribution of the volume change rate provides a clear representation of the impact of geometric deformation on model volume while also highlighting a few anomalous cases. These findings contribute to a deeper understanding of the relationship between geometric deformation and volume change, offering valuable insights for further optimization of deformation strategies and related studies.

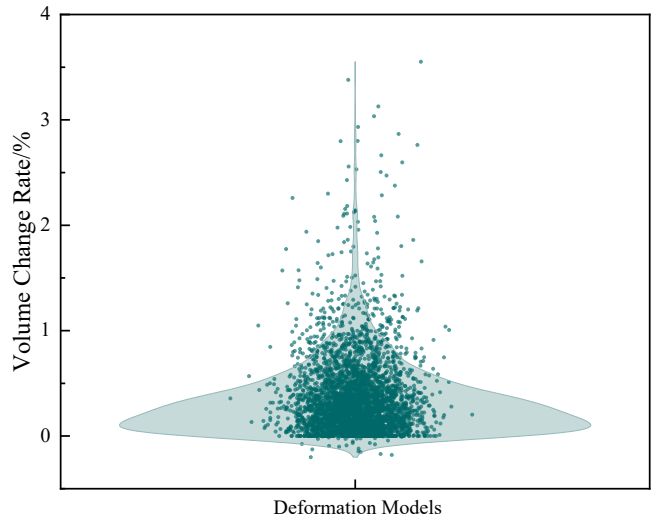

Fig.2 Distribution of volume change rate in deformation models.

To further validate the reliability of the CFD data, Fig. 3 presents the distribution of the maximum velocity ($V_{max}$) across all cases under different mass flow rates. The maximum velocity, calculated as $V_{max} = \max(\sqrt{u^2 + v^2 + w^2})$, shows a clear pattern in which most of the values are concentrated between 0 m/s and 1 m/s. This suggests that despite the increase in the inlet flow rate, the velocities generally remain within a stable and expected range. As the flow rate increases, the spread of the velocity distribution widens, indicating that the flow field becomes more dynamic and complex, with greater fluctuations in velocity. The shift from a narrow to a broader distribution reflects the increased variability of flow characteristics as mass flow increases. Similarly, Figure 4 illustrates the distribution of normalized pressure differences ($\Delta P^*$) for all cases under different flow conditions. The normalized pressure is calculated as $P^* = P / (0.5 * \rho * V_{max})$, where $P$ is the pressure, $\rho$ is the fluid density. The normalized pressure difference is calculated as the difference between the maximum and minimum values of the normalized pressure in the flow field. The normalization process removes the direct influence of flow and velocity on the absolute pressure values and makes the relative changes in pressure clearer. The results show that in most cases the range of normalized pressure differences is mainly concentrated between 0 and 50, and the density of the distribution is more concentrated in the lower range, suggesting that the flow is more stable in this range. This feature of the normalized pressure distribution provides an intuitive basis for understanding the flow field dynamics and further validates the reliability and accuracy of the CFD simulation results.

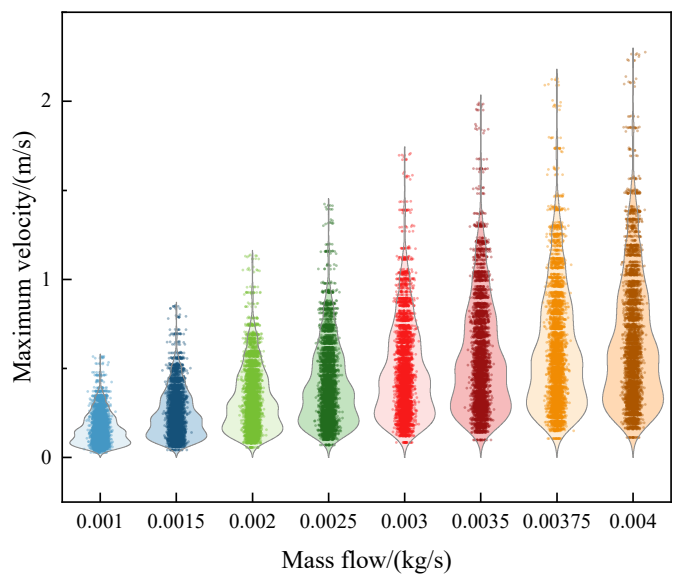

Fig.3 Maximum velocity distribution at different mass flow rates.

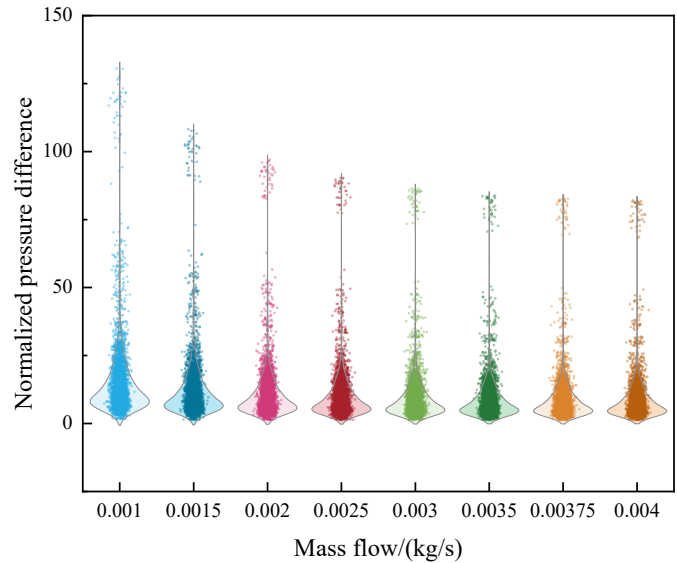

Fig.4 Normalized differential pressure distribution at different mass flow rates.

**3D Model Deformation**

To expand the dataset and generate additional synthetic aneurysm cases, de-aneurysm and deformation operations were performed on the original 3D models to simulate various aneurysm morphologies. Initially, the 3D model of the real aneurysm was imported into Geomagic Wrap 2021 software (3D Systems Inc., USA), where the Mesh Doctor tool was employed to identify and repair geometric issues, such as non-manifold edges, self-intersections, cusps, small tunnels, and holes, ensuring topological integrity and physical usability. The aneurysms were then removed from the original model, followed by the application of Ramesh operations. The Fit Surface function was subsequently used to generate NURBS surfaces to smooth and optimize the model, ensuring accurate geometry reconstruction and significantly enhancing surface smoothness and precision. The optimized model was saved in STEP format for subsequent boundary condition definition and numerical simulations.

Following aneurysm removal, a deformation operation was applied to the vascular region where the aneurysm had been located, simulating aneurysms with different morphologies by randomizing the size and shape of the model. Deformation amplitudes were randomly selected within the range [0.5, 1.0] to ensure the diversity of each synthesized case while maintaining physiological plausibility. After completing the deformation, Ramesh and Fit Surface optimizations were again performed to ensure the quality and accuracy of the models. Finally, these deformed models were also saved in STEP format. In addition to the aneurysm cases, the dataset also includes vessel data without aneurysms for comparative analysis. These deformed models are used not only to generate additional CFD simulation data but also to provide a broader range of case data to support in-depth analyses of aneurysm morphology, blood flow characteristics, and rupture mechanisms.

**3D Model Conversion to Segmentation Mask**

To facilitate the integration of hemodynamic simulations with medical imaging data, the above 3D models were converted into segmentation mask files resembling medical imaging data using 3D Slicer software. These generated mask files play a critical role in aligning hemodynamic data with imaging data, ensuring an accurate representation of the aneurysm geometry. This conversion process not only improves the spatial consistency and accuracy of the model but also enhances the precision and reliability of hemodynamic analysis. By directly aligning the results of computational fluid dynamics (CFD) simulations with the imaging data, this alignment ensures that the CFD results are accurately represented within the context of the patient's specific anatomy. Additionally, this method strengthens the correlation between CFD results and hemodynamic parameters, providing substantial support for future studies that combine imaging data and CFD results for AI-based modeling. The integration of image data with hemodynamic simulations plays a vital role in improving diagnostic accuracy and advancing personalized treatment strategies.

**Mesh Generation**

The processed STEP file was imported into SpaceClaim software (ANSYS Inc., USA), where inlet, outlet, wall, and fluid regions were defined. Since the AneuX[12] dataset lacks imaging data, the cross-section with the largest pipe diameter was selected as the inlet surface, which was subsequently verified by an expert in computational fluid dynamics. Once the boundary conditions were defined, the model was saved in SCDOC format for the mesh generation process. During the mesh generation phase, Fluent

Meshing 2023 R1 software (ANSYS Inc., USA) was used to create unstructured polyhedral meshes for the fluid domain of the intracranial aneurysm. Polyhedral meshes were preferred over traditional hexahedral or tetrahedral meshes due to their superior accuracy and computational efficiency. These meshes offer performance close to hexahedral meshes, but with significantly better computational efficiency compared to tetrahedral meshes. Additionally, polyhedral meshes are highly effective in handling complex geometries and boundary layer details[20]. The minimum mesh size was set to 0.15 mm, with the near-wall boundary layer mesh divided into 10 layers, each exhibiting a growth rate of 1.2[21]. This configuration enhanced the accuracy of the calculations, particularly when analyzing wall shear stress (WSS).

### Boundary Condition Definition and Hemodynamic Simulation

In this study, blood was modeled as an incompressible Newtonian fluid with a density of 1050 kg/m³ and a dynamic viscosity of 0.00345 Pa·s to represent the laminar flow behavior within an intracranial aneurysm. The blood vessel wall, inlet, and outlet were defined as rigid no-slip, mass flow inlet, and zero-pressure outlet boundary conditions, respectively. Mass flow rates for steady-state simulations ranged from 0.0010 to 0.0040 kg/s and included values of 0.0010 kg/s, 0.0015 kg/s, 0.0020 kg/s, 0.0025 kg/s, 0.0030 kg/s, 0.003 kg/s, 0.00375 kg/s, and 0.0040 kg/s. The OpenFOAM was used as the computational tool, and the finite volume method (FVM) was employed to solve the three-dimensional Navier-Stokes equations for incompressible Newtonian fluids. The icoFoam solver was applied for steady-state simulations. To ensure accurate pressure-velocity coupling, the PISO (Pressure Implicit with Splitting of Operators) algorithm was implemented for steady simulations.

The simulation results yielded key hemodynamic parameters, including velocity ($u$, $v$, $w$), pressure ($p$), and wall shear stress, which are essential for characterizing the fluid dynamics within intracranial aneurysms. These parameters provide critical insights into the hemodynamic environment and serve as important indicators of aneurysm formation, progression, and rupture risk. By analyzing these parameters, regions of high-risk flow behavior can be identified, which are crucial for assessing aneurysm stability. Moreover, these results have significant clinical implications, including risk assessment, treatment planning, and the development of targeted therapeutic strategies aimed at mitigating rupture risk and improving patient outcomes.

## Technical Validation

The dataset generation process underwent strict quality control and validation to ensure its scientific validity and effectiveness. Firstly, all 3D models in the dataset were based on real brain aneurysm data, and synthetic data were generated through controlled deformation to ensure diversity in aneurysm morphology and anatomical consistency.

Secondly, to ensure that the simulation results are not influenced by the mesh size, a sensitivity analysis of multiple mesh sizes was conducted. As shown in Fig. 5, the minimum grid sizes of 0.05 mm, 0.10 mm, 0.15 mm, 0.20 mm, and 0.25 mm correspond to total grid numbers of approximately 150,000, 250,000, 330,000, 330,000, and 880,000, respectively. The analysis indicates that the accuracy of numerical calculations improves as the grid size decreases. However, when the minimum mesh size is reduced to 0.15 mm, the computational results closely align with those obtained using a grid size of 0.10 mm, with a relative error of less than 1%. Therefore, to balance computational accuracy, efficiency, and resource consumption, a minimum grid size of 0.15 mm was ultimately selected for this study.

In addition, all CFD calculations for the cases were performed with the Courant–Friedrichs–Lewy (CFL) number consistently maintained below 1. This is crucial for ensuring the stability of the numerical method and the reliability of the results. Keeping the CFL number below 1 effectively controls the stability of the numerical calculations, preventing issues such as instability or non-convergence that could arise from an excessively high CFL number. By ensuring that the CFL number remains within a reasonable range, the accuracy of fluid dynamics simulations is preserved, avoiding errors due to numerical instability.

Finally, the residual convergence curves for velocity ($u$, $v$, $w$) and pressure ($p$) were analyzed to evaluate the reliability of the simulation results. As shown in Fig.6, the residuals of the velocity components ($u$, $v$, $w$) decrease rapidly from $10^{-3}$ to approximately $10^{-9}$ within the first 20,000 iterations and eventually stabilize around $10^{-9}$, demonstrating high accuracy and strong convergence for the velocity field. In contrast, the pressure residual ($p$) decreases gradually, accompanied by slight fluctuations in the initial stage, from $10^{-3}$ to below $10^{-5}$. Although the pressure residual does not achieve the same level of accuracy as the velocity components, it stabilizes over time, indicating that the converged results exhibit acceptable accuracy. These findings confirm the suitability of the selected mesh size and ensure the accuracy and computational efficiency of the simulation.

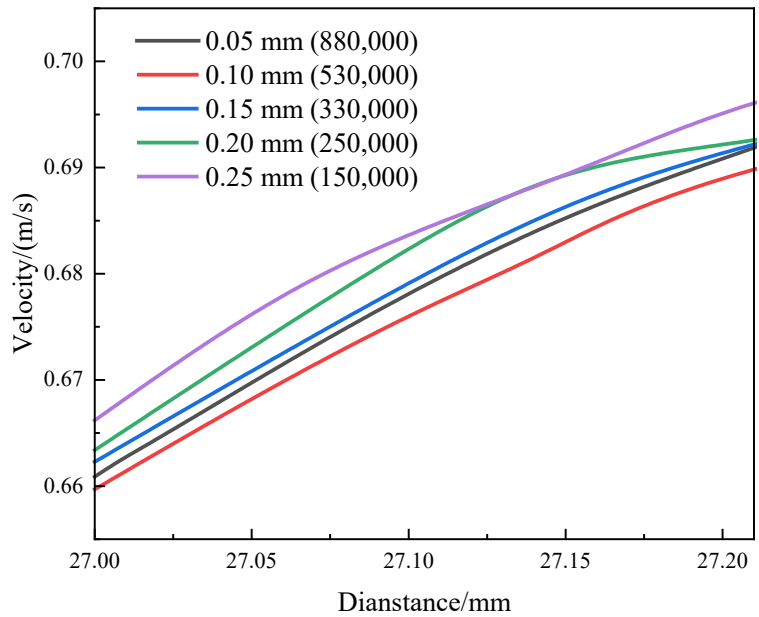

Fig.5 Grid independence analysis.

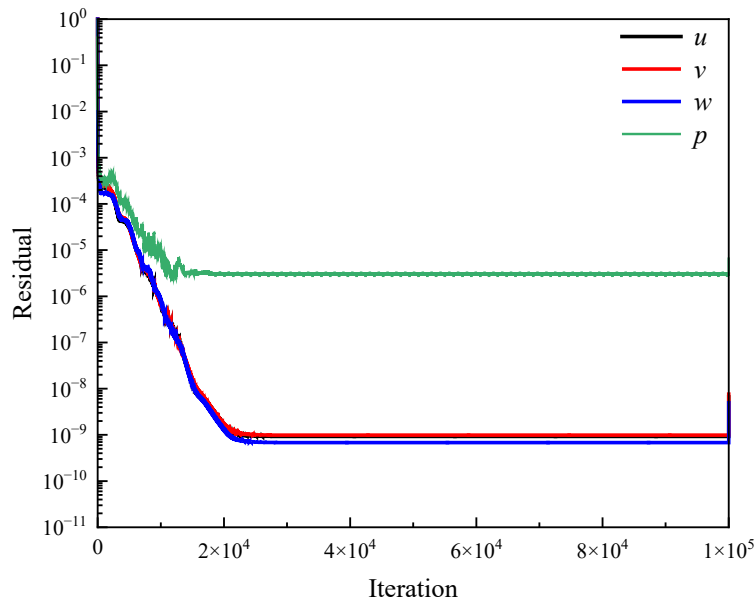

Fig.6 Residual convergence curve of velocity and pressure.

**Data Records**

The dataset is available at https://github.com/Xigui-Li/Aneumo. The dataset comprises 468 real aneurysm models and 10,000 synthetic aneurysm models, with synthetic models including 466 control models (without aneurysms) and 9,534 deformed aneurysm models. It provides high-quality 3D STL files, mesh files for computational fluid dynamics (CFD) simulations, and mask files for both aneurysms and their corresponding arteries. These resources are invaluable for basic research and CFD simulations of intracranial aneurysms. Additionally, the dataset includes hemodynamic parameters across eight different steady-state conditions, covering essential metrics such as velocity, pressure, and wall shear stress. The file formats include mask files (NIfTI), 3D model files (STL), mesh files (MESH), and hemodynamic parameter files (VTK). The dataset is particularly suitable for developing new computational models, validating clinical prediction tools and conducting more in-depth scientific studies.

**Usage Notes**

Users should cite this paper in their research output and acknowledge the contribution of this dataset in their study.

**Code availability**

No custom code was utilized in this study. Software tools used for data processing are mentioned in the Methods section.

**Acknowledgments**

This paper was supported by the National Natural Science Foundation of China (Grant No. 82394432 and 92249302), the Shanghai Municipal Science and Technology Major Project (Grant No. 2023SHZDZX02).
The computations in this research were performed using the CFFF platform of Fudan University.

**Author Contributions**

Xigui Li: Conceptualization, Methodology, Validation, Formal analysis, Writing - original draft, Writing - review & editing. Yuanye Zhou: Conceptualization, Methodology, Software, Validation, Review & editing. Feiyang Xiao: Conceptualization, Methodology, Investigation, Software. Xin Guo: Methodology, Investigation, Review & editing, Supervision, Resources, Project administration. Yichi Zhang: Review & editing. Chen Jiang: Review & editing, Supervision, Project administration. Jianchao Ge: Resources, Data Curation. Xiansheng Wang: Resources, Data Curation. Qimeng Wang: Resources, Data Curation. Taiwei Zhang: Resources, Data Curation. Chensen Lin: Methodology, Supervision. Yuan Cheng: Supervision, Resources, Project administration, Funding acquisition. Yuan Qi: Supervision, Project administration.

## Competing interests

The authors declare that they have no known competing financial interests or personal relationships that could have appeared to influence the work reported in this paper.